%% file: main.tex
\documentclass{article}
\usepackage{graphicx}
\usepackage{amssymb}
\usepackage{amsmath}
\usepackage{mathtools}
\usepackage{algorithmic}
\usepackage{booktabs} 
\usepackage{esvect}
\usepackage{enumitem}
\usepackage[ruled]{algorithm2e}

\usepackage[final]{corl_2019} 

\title{Active 6D Multi-Object Pose Estimation in Cluttered Scenarios with Deep Reinforcement Learning}

\author{
  Juil Sock, Guillermo Garcia-Hernando, Tae-Kyun Kim \\
  Imperial College London, UK\\
  }

\begin{document}
\maketitle


\begin{abstract}
In this work, we explore how a strategic selection of camera movements can facilitate the task of 6D multi-object pose estimation in cluttered scenarios while respecting real-world constraints important in robotics and augmented reality applications, such as time and distance travelled. In the proposed framework, a set of multiple object hypotheses is given to an agent, which is inferred by an object pose estimator and subsequently spatio-temporally selected by a fusion function that makes use of a verification score that circumvents the need of ground-truth annotations. The agent reasons about these hypotheses, directing its attention to the object which it is most uncertain about, moving the camera towards such an object. Unlike previous works that propose short-sighted policies, our agent is trained in simulated scenarios using reinforcement learning, attempting to learn the camera moves that produce the most accurate object poses hypotheses for a given temporal and spatial budget, without the need of viewpoints rendering during inference. Our experiments show that the proposed approach successfully estimates the 6D object pose of a stack of objects in both challenging cluttered synthetic and real scenarios, showing superior performance compared to strong baselines.
\end{abstract}



\section{Introduction}
	
\begin{figure}[t]
\centering
\includegraphics[width=2\linewidth]{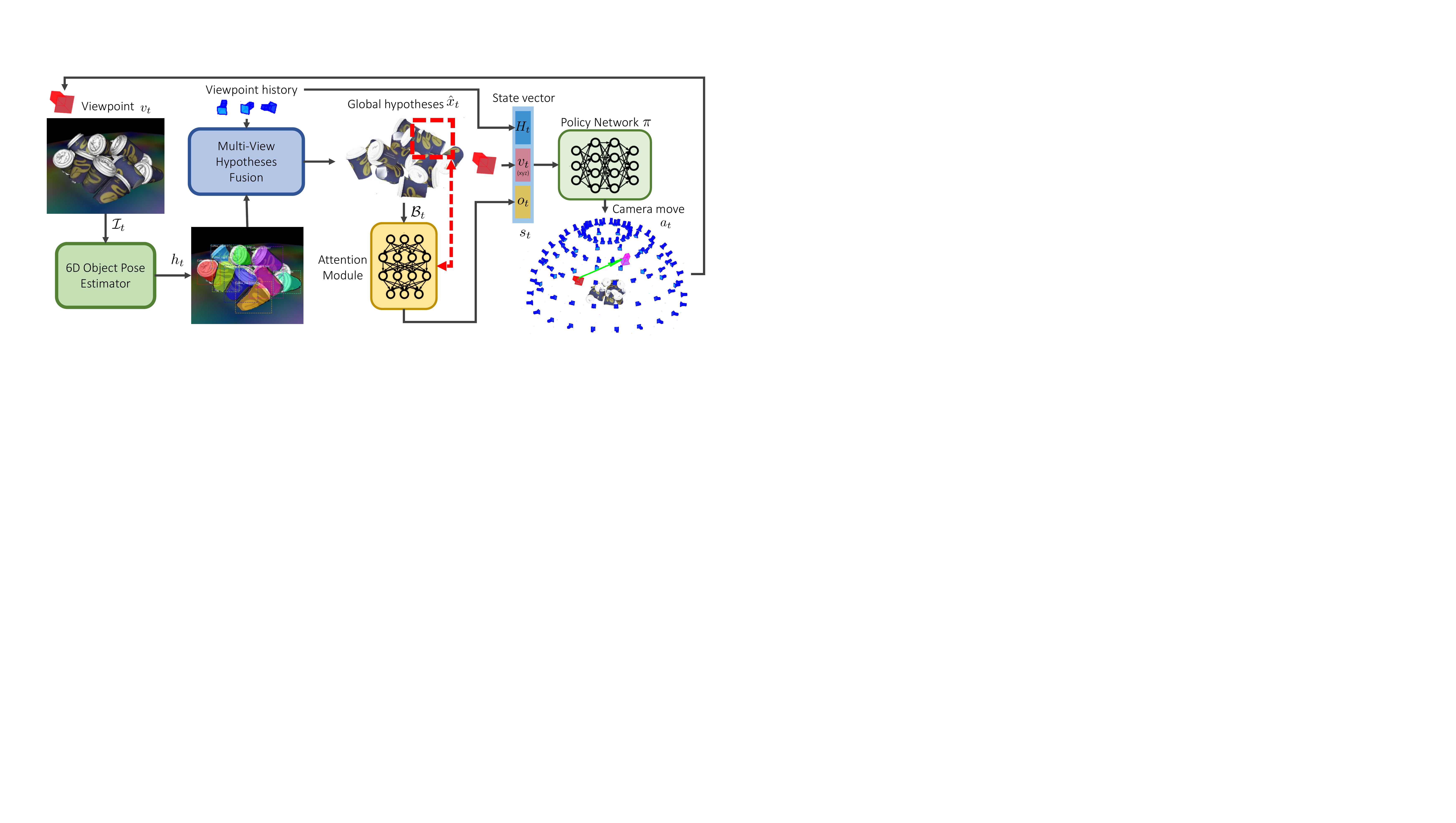}
\vspace{-27em}
\caption{\textbf{Proposed framework}. An image from viewpoint $v_t$ is passed through a 6D object pose estimator which generates pose hypotheses. Current and previous hypotheses are accumulated using a fusion function. The agent analyzes the scene, focusing its attention to the object which it is most uncertain about and moving the camera towards such object.}
\vspace{-1.5em}
\label{fig:overview}
\end{figure}

Accurate 6D object pose estimation of multiple objects on a cluttered scenario may become essential in applications such as robotic manipulation and augmented reality which, currently lacking proper solutions, resort to either coarse approximations~\cite{izadi2011kinectfusion} or intermediate and less interpretable solutions~\cite{levine2016end}. Such applications and scenarios are naturally framed in an active setting, where either a robot or a human has the capability of moving the camera to different viewpoints as presented in e.g. the Amazon Picking Challenge~\cite{eppner2016lessons,zeng2017multi}. 
The present work explores how moving the camera in the scene, reaching different viewpoints, can help to overcome the inherent challenges in 6D object pose estimation, such as clutter and occlusion, identifying two ends of the spectrum of such problem. On one end we can decide not moving the camera at all, reducing the problem to single shot object pose estimation and thus subject to the object pose estimator limitations \cite{hinterstoisser2011linemod, brachmann2014learning, tejani2014latent, hodan2018bop}. On the other end, the camera can be moved to cover all the possible viewpoints~\cite{zeng2017multi}, however, this does not respect real-world constraints, such as time limitation in terms of a number of movements and energy expenditure in terms of distance traveled. We are interested in finding a compromised solution that moves the camera strategically by reaching the most informative viewpoints given a limited number of camera movement and keeping the traveled distance as low as possible.
Previous work in active 6D object pose estimation includes the work of Doumanoglou \textit{et al.} \cite{doumanoglou2016recovering}, which explored the problem of estimating the next best camera viewpoint by aiming to reduce the uncertainty of the pose estimator in terms of entropy reduction. Sock \textit{et al.}~\cite{sock2017multi} take a similar entropy reduction approach, but instead of using pose inference, they propose a heuristic geometric approach to estimate view entropy. Both \cite{doumanoglou2016recovering} and \cite{sock2017multi} present two severe limitations. First, both frameworks are short-sighted, meaning that they move the camera to the next best view independently of the past and future viewpoints and potentially reaching a redundant view. Second, they require rendering different views before making a camera movement, which is an important limitation on real-world applications where the number of possible viewpoints is high or even infinite. 
This work proposes a framework to tackle the active 6D object pose estimation problem that overcomes the limitations of the previous work and it is depicted in Fig. \ref{fig:overview}. Given a set of objects hypotheses inferred by an object pose estimator, our framework first fuses the information taking into account the previous estimations and chooses the hypotheses that best explain the scene without needing ground-truth information with the use of a verification score. The result of this fusion function is given to an agent, which analyzes all the hypotheses. Using an attention mechanism, the agent directs its attention to the object which it is most uncertain about and moving the camera towards such an object. The agent is trained in simulated scenarios on a reinforcement learning framework, attempting to learn a policy that moves the camera producing the most accurate object poses hypotheses for a given temporal and spatial budget. The sequential nature of the reinforcement learning framework gives the agent the ability to reason temporally and make long term decisions based on previous movements. Furthermore, our careful state space design allows the agent to perform inference without the need for rendering the entire amount of viewpoints at each time step. We evaluate our framework on challenging synthetic and real scenarios, showing that our framework can produce camera movements that achieve robust 6D object pose estimation over strong baselines.  
In summary, this paper contains the following main contributions: 
\begin{enumerate}
    \item An active 6D multi-object pose estimation within a reinforcement learning framework is proposed. State and action spaces are carefully designed and a reward function is tailored to the problem of interest. 
    \item Application specific modules are proposed: a hypotheses fusion function that works without ground-truth annotations by a proposed verification score and an object-specific attention module that directs the attention of the agent.
    \item Extensive experimental evaluation of both the proposed method and baselines on both synthetic and real cluttered scenarios.
\end{enumerate}


\section{Related Work}
\label{sec:related_work}

\noindent\textbf{6D object pose estimation.} 6D object pose estimation from a single image has been extensively researched in the past decades~\cite{sahin2019instance}. Accurate pose estimation can be obtained under moderate occlusion and clutter with handcrafted features~\cite{drost2010model, hinterstoisser2011linemod}. To handle occlusion and truncation, keypoint detection with PnP~\cite{Rad_2017_ICCV,tekin2017real} or per-pixel regression/patch-based approaches~\cite{brachmann2014learning,doumanoglou2016recovering,tejani2014latent} followed by Hough voting~\cite{peng2018pvnet,tejani2014latent} or RANSAC have been proposed. Most recent approaches attempt to jointly learn features and pose estimation using neural networks~\cite{Balntas_2017_ICCV} in either RGB~\cite{Kehl_2017_ICCV,tekin2017real,xiang2017posecnn} or RGB-D images~\cite{sock18multi,wang2019normalized}. However, many of the methods are designed and evaluated for Single instance of single object~\cite{hodan2018bop} and pose estimation of multiple instances of single object in cluttered scenarios remains a challenge. Estimating object poses on such scenarios using the above methods is challenging due to cascading errors due to severe occlusions and foreground clutter. Sock \textit{et al.}~\cite{sock18multi} presented a work specifically for such scenarios by generating training dataset with occlusion patterns. Sundermeyer \textit{et al.}~\cite{sundermeyer2018implicit} learned an implicit representation of object orientations defined by samples in a latent space to handle occlusions. However, the accuracy of a pose hypothesis is fundamentally limited by the visibility of the object in the input image. We end this section by reviewing methods that use multiple view information~\cite{viksten2006increasing,collet2010efficient,erkent2016integration,zeng2017multi,kanezaki2018rotationnet,li2018unified}. However, these methods assume a set of images captured from pre-determined viewpoints are available. Our framework is agnostic to the choice of object pose estimator and we propose a fusion function to incorporate multi-view information.

\noindent\textbf{Active vision, object detection and poses.} We review recent work on the active vision that aims to either improve the detection of objects, their pose estimation or both. Several methods have been proposed to select glimpse on static image to accelerate object detection~\cite{mathe2016reinforcement,gonzalez2015active,caicedo2015active} and more recently view selection for a moving observers such as visual navigation~\cite{zhu2017target,mirowski2016learning,gupta2017cognitive} and classification~\cite{calli2018active}. In \cite{ammirato2017dataset} an active vision dataset and a reinforcement learning based baseline to explore the environment to detect objects are proposed.  \cite{cheng2018reinforcement} presents a method to jointly learn a policy for both grasping and viewing. The method uses a simulated environment which detects objects depending on the occlusion rate. Unlike our framework where multiple objects are present, both methods include one object per scene. \cite{cheng2018geometry} shows a geometry-aware neural networks which integrate different views to a latent feature tensor which is also used to select views for the purpose of object reconstruction and segmentation. Similarly ~\cite{yang2018active} uses reinforcement learning framework to select views to reconstruct 3D volume from RGB images of single object.  More related to our work, \cite{doumanoglou2016recovering} proposes a Hough forest approach and an entropy-based viewpoint selection. An important drawback of this method is that it does not have a mechanism to resolve conflicting next best view selection in the presence of multiple hypotheses. ~\cite{sock2017multi} integrates different components to build a complete active system which detects and pose estimates multiple objects. However, both frameworks decide the next best view independently of the past and future decision and also requires rendering of different viewpoints.

\noindent\textbf{Attention to objects.} Recent work has shown that employing attention mechanism is useful in the presence of multiple objects, distractors, and clutter. ~\cite{xu20163d} used attention to select discriminative features for the purpose of view selection for object classification given a single detected object, whereas we use attention module to select a hypothesis from a set of detected instances. ~\cite{kim2017interpretable} uses a pixel-wise attention module for the self-driving system to build more interpretable agent. \cite{devin2018deep} presents a system for object grasping with object-level attention where an attention mechanism is used to select the object to be manipulated on specific tasks. Similarly, in the proposed framework we propose an attention mechanism that makes the agent focus on the most uncertain object.

\section{Proposed Framework}
\label{sec:proposed_framework}
\noindent\textbf{Problem definition.} Given a target stack of $n$ objects, our aim is to infer the 6D pose $x^i\in \mathbb{R}^6$ of every $i$ visible object in the stack, consisting of its 3D location and 3D orientation in the scene. In other words, our objective is to find the set of object hypotheses $\hat{x}=\{\hat{x}^1,...,\hat{x}^N\}$ that best describes the scene. We formulate the problem in an active setting, where an agent has the ability to navigate through a finite set of viewpoints $\mathcal{V} = \{v^1,...v^M\}$, where $v^i \in \mathbb{R}^6$ is the 6D camera pose, and $M=|\mathcal{V}|$ is the number of viewpoints accessible to the agent. At each time step $t$, the agent observes the state of the scene $s_t$ from a viewpoint $v_t \in \mathcal{V}$ and decides an action $a_t$ following a policy function $\pi$ that moves the agent to next viewpoint $v_{t+1}$. The agents proceeds until it reaches the maximum episode length $T$ generating a trajectory $\tau = (s_0, a_0, ..., s_{T-1}, a_{T-1})$. Moving an agent in the real world is costly in terms of time $T$ and energy consumption, which we indirectly measure as a function of total traveled distance $d$. Both $d$ and $T$ are budget-constrained and low values are desired.

\noindent\textbf{Framework overview.} In our framework, presented in Fig. \ref{fig:overview}, at each time step $t$ an input image $\mathcal{I}_t$ is acquired from a viewpoint $v_t$ and it is passed through a 6D object pose estimator that provides pose hypotheses $h_t$ on the currently observed objects. These hypotheses are accumulated using a fusion function that considers both the current hypotheses and the previously observed ones and a selection of such is given to the agent. In a cluttered scenario, there are multiple objects with different challenges such as occlusion or measuring pose estimation confidence. This indicates there may not be a single next view ideal for all the objects and thus the agent reasons about the scene and focus its attention to the object which it is most uncertain about and moves the camera towards such object. This process is repeated until the maximum fixed number of time steps $T$ and then the 6D pose hypotheses for the scene $x$ is obtained. In the following sections, we describe the different components of our framework.

\subsection{6D object pose estimation}
Given an image $\mathcal{I}_t$ acquired from viewpoint $v_t$, the 6D object pose estimator outputs a set of object hypotheses for the current view $h_t = \{h^1_t, ..., h^k_t\}$, where $h^i_t \in \mathbb{R}^6$ and $k$ might differ from the actual number of objects in the scene $N$. Our framework is agnostic to the 6D object pose estimator of use and in this work, we opted to regress pixel-wise object-coordinate maps in a similar way to \cite{wang2019normalized}. We use a 6D object pose estimator working on the RGB-D domain which, in addition to object hypotheses, also provides 2D bounding boxes and segmentation masks for each detected object, which will be used on different stages of our framework detailed below. 

\subsection{Multi-view hypotheses fusion}
At each viewpoint, $v_t$, the agent obtains a different set of object hypotheses $h_t$ from the 6D object pose estimator. To obtain a global scene hypothesis $\hat{x}$ that accumulates the information obtained from different viewpoints, we define a hypotheses fusion function $f$. Given a history of object hypotheses $h_1, ..., h_t$ and camera viewpoints up to time step $t$, the function $f$ outputs the accumulated hypotheses $\hat{x}_t$ in the global coordinate system. Intuitively, $f$ gathers all the object hypotheses that have been observed so far, selecting the ones that better explain the scene among all the pool of hypotheses, hence $\hat{x}_t \subseteq \bigcup_{i=1}^{t} \bigcup_{j} h^j_i$. In our framework, $f$ makes use of hierarchical clustering to group hypotheses and the hypothesis with the highest object pose confidence score is selected from each cluster in a similar way to \cite{sock2017multi}. In our framework, the hypothesis with the highest verification score from each cluster is chosen, as we use verification score to represent quality of the hypothesis.

\textbf{Verification score.} The hypotheses fusion function requires quantifying the quality of a hypothesis in the absence of ground-truth information. For this, we introduce a verification score that estimates hypothesis confidence inspired by the work of \cite{aldoma2012global} and \cite{doumanoglou2016recovering}. These works assume that a scene point $q$ that belongs to an object -is an \textit{inlier}- if the distance to the nearest neighboring point $p$ of a rendered 3D model of a hypothesis is lower than a certain threshold $\epsilon$. This assumption is not generally true in the presence of multiple objects in clutter. To overcome this limitation, we use the segmentation mask from the object pose estimator and consider points on the depth map that lie inside the mask as inliers.
The local fitting between $p$ and $q$ is measured by $\delta(p,q)$ and is defined as follows:

\begin{equation}
\label{eqn:verification_score}
  \delta(p,q)=
  \begin{cases}
    \frac{1}{2}(1-\frac{\|p-q\|_2}{\epsilon})+\frac{1}{2}{(n_p \cdot n_q)}, & \text{if } \|p-q\|_2<\epsilon\\
    0, & \text{otherwise}
  \end{cases},
\end{equation}

where $n_p$ and $n_q$ denote the normal vectors at $p$ and $q$ respectively. The verification score $c$ is calculated as the mean value of $\delta(p,q)$ for all the inlier points. In the supplementary, we show how this score correlates with the object pose error making it suitable in the absence of ground-truth annotations.

\textbf{Objects feature representation.} When encoding object features, 6D object pose hypothesis information is not directly used given that some estimated object poses are of poor quality to provide useful enough information. Instead, we propose to use a more reliable and simpler feature representation to encode object hypotheses. The output of the hypotheses selector is a set of object features $\mathcal{B}_t$ for $k$ object hypotheses. Each element in $\mathcal{B}_t$  consists of a tuple $(b^i, d^i, c^i)$, where $b^i$ is the normalized 4 dimensional object 2D bounding box coordinates, $d^i$ is the distance of the hypothesis from the viewpoint in depth values on the camera coordinates and $c^i$ is the verification score of the hypothesis. Bounding box coordinates are directly used for $b^i$ and $d^i$ is approximated by averaging the mean value of depth map values corresponding to the segmentation mask of the 6D object pose estimator. The value from the object hypothesis in depth axis is not directly used as it is often incorrect notably when the object is highly occluded.

\subsection{Object attention module}
\label{sec:attention}

At a given time step $t$, the agent observes a number of object hypotheses and makes a choice on which next viewpoint to visit at $t+1$. To guide the agent on that decision, we propose an attention mechanism that makes the agent to focus on the most uncertain object and move accordingly. In this section, we describe how we design this attention mechanism and in Sec. \ref{sec: policy} we provide details on its learning function.

Our attention mechanism receives as input the set of object features $\mathcal{B}_t$ from the previous module and returns an individual object $o_t$ feature of the object to reason about and its index $m_t$ in a similar way to ~\cite{devin2018deep} and \cite{wang2018deep}. First, a representation of each object in $\mathcal{B}_t$ is extracted using a fully connected network. These individual features are aggregated using a mean-pool layer leading to a global object representation that is concatenated to the individual feature. A Selector network takes these concatenated features and outputs a scalar score indicating the importance of the object which is normalized using a softmax layer. The object with the highest score is then selected and its features are given to the policy network. Note that gradients flow from the policy network, guiding the attention module to the object of interest in our active setting. A description of the algorithm can be found in Alg. \ref{algo:attention}.%

\begin{algorithm}[H]
  \KwData{$\mathcal{B}_t$ features of $k$ objects at time step $t$}
  \KwResult{$o_t$ object feature and object index $m_t$ for the policy network}
 \For{$b^i \in \mathcal{B}_t$}{
  $g^i := FC(b^i,d^i,c^i)$  \tcp{fully connected network}
  }
  $g^{global} := meanpool(g^1,...g^k)$ \\
  \For{$b^i \in \mathcal{B}_t$}{
  $w^i := Selector(concat(g^i,g^{global}))$  \tcp{object score}
  }
  $\bar{w}^1,...\bar{w}^k = softmax(w^1,...w^k)$ \\
  $m_t = argmax(\bar{w}^1,...\bar{w}^k)$ \tcp{attended object index}
  $o_t = concat(g^m * \bar{w}^m, b^m, d^m, c^m)$ \tcp{individual object feature}

 \caption{Attention module}
 \label{algo:attention}
\end{algorithm}

\subsection{Policy learning}
\label{sec: policy}
\textbf{Formulation of reinforcement learning problem.} We formulate the decision process of choosing an action $a_t$ leading to a viewpoint $v_{t+1}$ within a reinforcement learning (RL) framework. The next viewpoint is strategically chosen to lead to an optimal set of hypotheses $\hat{x}^*$ that best describes the scene. The optimization problem is defined as the maximization of a parametric function $J(\theta)$ defined as 
$J(\theta) = \mathbb{E}_{\tau \sim p_\theta(\theta)} \left[\sum_{t = 0}^T \gamma^t r_t \right]$, where $ p_\theta(\theta)$ is the distribution over all possible trajectories following the parametric policy $\pi_\theta$. The term $\sum_{t = 0}^T \gamma^t r_t$ represents the total return of a trajectory for a horizon of $T$ time steps and a discount factor $\gamma \in [0,1]$. 

\noindent\textbf{Reward function.} 
Our reward function involves three different terms: (i) individual pose estimation accuracy; (ii) change in distance from camera to the attended object and (iii) penalization for long distance camera movements. The first term is defined as follows:
\begin{align}
\label{eqn:score_reward}
\begin{split}
r^{e_{ADD}}_t = e_{ADD}(\hat{x}^m_{t+1},x^m_{t+1}) - e_{ADD}(\hat{x}^m_{t},x^m_{t}),
\end{split}
\end{align}
where index $m_t$ denotes the object index selected by attention network described in sec.\ref{sec:attention} at time $t$.  $e_{ADD}$ is the most widely used 6D object pose error function in the literature and is the average Euclidean distance of model points proposed by Hinterstoisser \textit{et al.}~\cite{hinterstoisser2012model}. It is defined as the average Euclidean distance of the estimated pose $\hat{x}$ with respect to the ground-truth pose $x$ of an object model. 

We hypothesize that a viewpoint closer to the object of interest is more likely to give better pose estimation. It can be used to guide and accelerate the policy learning and this leads to the second term defined as:
\begin{align}
\label{eqn:obj_dist_reward}
\begin{split}
r^{dist}_t =\|\hat{x}^m_{t+1}[\vv{xyz}]-v_{t+1}[\vv{xyz}]\|_2 - \|\hat{x}^m_{t}[\vv{xyz}]-v_{t}[\vv{xyz}]\|_2,
\end{split}
\end{align}
where $[\vv{xyz}]$ is the translation vector from a 6D pose. The full reward function is the combination of the two above and another term penalizing long distance camera motions aiming to minimize the total distance traveled $d$:
\begin{align}
\label{eqn:reward_function}
\begin{split}
r_t = (1-\alpha) r^{e_{ADD}}_t + \alpha\beta r^{dist}_t - (1-\alpha)(1-\beta)(\|v_{t+1}[\vv{xyz}]-v_{t}[\vv{xyz}]\|_1).
\end{split}
\end{align}
where $\alpha$ and $\beta$ are hyperparameters in $[0,1]$ weighting the influence of the different reward terms.

\noindent\textbf{State space.} The state vector $s_t$ in our framework consists of three concatenated components. First, the object feature $o_t$ that comes from the attention module presented in Algorithm~\ref{algo:attention}. Second, the current camera pose position $v_{t}[\vv{xyz}$]. Last, a history vector $H_t$ that encodes previous camera positions for $T$ steps, which is initialized to zeros and it is filled at each time step. This history vector is needed in the absence of a memory module to respect the Markov property on a reinforcement learning framework.

\noindent\textbf{Action space.} We define the action space as a two-dimensional continuous space spanning azimuth and elevation angles the camera viewpoint of a hemisphere centered in the stack of objects. More specifically, an action at time step $t$ is defined as $a_t=\{\phi^a_t, \phi^e_t\}$ where $\phi^a$ and $\phi^e$ represent azimuth and elevation angles of the camera respectively. Given the finite nature of the viewpoint space, the angles from $a_t$ are mapped to the closest viewpoint in $\mathcal{V}$. 

\noindent\textbf{Policy network}
We represent the policy function $\pi_\theta$ as a neural network with parameters $\theta$ that outputs the mean and standard deviation of a Gaussian distribution. To optimize $\theta$ different methods can be used, however, in our framework, we use policy gradients method~\cite{schulman2017proximal}. These methods optimize $J(\theta)$, where the gradient of the expected return $\nabla_\theta J(\theta)$ is estimated with trajectories sampled by following the policy. Details of the architecture are discussed in the supplementary material.
\section{Experimental Results}
\label{sec:result}
\begin{table}[b]
\caption{Detection rate, $e_{ADD}$ and distance travelled evaluated with the different baselines on synthetic environment. Maximum episode length set to 5.}
\vspace{1em}
\label{tab:main_results}
\centering
\resizebox{0.85\textwidth}{!}{
\begin{tabular}{@{}lcccccc@{}}
\toprule
 & \multicolumn{3}{c}{Coffee cup} & \multicolumn{3}{c}{Bunny} \\ \cmidrule(l{10pt}r{10pt}){2-4} \cmidrule(l{10pt}r{10pt}){5-7} 
Policy & \begin{tabular}[c]{@{}c@{}}Distance\\ $d$ $\downarrow$\end{tabular} & \begin{tabular}[c]{@{}c@{}} $e_{ADD}$\\ (mm) $\downarrow$ \end{tabular} & \begin{tabular}[c]{@{}c@{}}Detection \\ rate $\uparrow$\end{tabular} & \begin{tabular}[c]{@{}c@{}}Distance\\ $d$ $\downarrow$\end{tabular} & \begin{tabular}[c]{@{}c@{}}$e_{ADD}$\\ (mm) $\downarrow$\end{tabular} & \begin{tabular}[c]{@{}c@{}}Detection \\ rate $\uparrow$\end{tabular} \\ \midrule
Random & 4.97 & 13.48 & 0.76 & 3.97 & 26.69 & 0.19 \\
Maximum Distance & 7.66 & 14.02 & 0.75 & 7.68 & 26.83 & \textbf{0.46} \\
Unidirectional & \textbf{2.88} & 14.32 & 0.74 & 2.88 & 27.12 & 0.44 \\\midrule
\textit{Proposed} & 3.71 &\textbf{ 11.12} & \textbf{0.80 }& \textbf{2.43} & \textbf{25.35} &\textbf{ 0.46} \\ \bottomrule
\end{tabular}
}
\end{table}

\noindent\textbf{Dataset and experimental setup.} For synthetic experiments, Bin-picking dataset with two different object models are used for training and testing: Coffee Cup from ~\cite{doumanoglou2016recovering} dataset and Bunny model from ~\cite{bregier2017iccv} dataset. For each model, we generate 40 scenes with different number of objects and random pose configurations. Objects with low visibility not detected in any viewpoints are removed from the ground-truth and are not included in the evaluation. For each object, 30 scenarios are used for training and 10 scenarios are used for testing. \\
We use the coffee cup scenario from the dataset of Doumanoglou \textit{et al.}~\cite{doumanoglou2016recovering} for the real experiment. Since the viewpoints of the dataset are not evenly distributed, a view grid is constructed in the same ways as the synthetic environment and the image from viewpoint closest to the grid viewpoint is used as the observation. During training and testing, 2D bounding box coordinate is rotated around the depth axis to correct the in-plane rotation. 
 The inference time for the pose estimator depends on the number of objects in the scene. In the case of 14 objects the pose estimator outputs hypotheses at 2 Hz. The rest of the pipeline including object accumulation, clustering and policy network operate at 25 Hz. It is possible design a pipeline to leverage intermediate image data acquired while travelling from $v_{t}$ to $v_{t+1}$, however it is less practical due to the processing time.\\
Our framework implementation is built in PyTorch and the learning algoritm, PPO \cite{schulman2017proximal}. More implementation details can be accessed in the supplementary material.\\

\noindent\textbf{Baseline policies.}
\label{sec:baselines} We test our active pose estimation system against a variety of baselines. The first baseline \textbf{``Random''} samples random actions from a fixed Gaussian distribution. The second baseline \textbf{``Unidirectional'' } moves the camera in one horizontal direction around the scene. The trajectory is generated such that it completes one full revolution around the scene. Elevation of the viewpoint is fixed to be 45 degrees which provide a balanced view between occlusion and variability in each view. The third baseline, \textbf{``Maximum distance''}, selects viewpoints with the longest distance from previous viewpoints. This baseline test whether the most informative views are simply far-apart views. Lastly, an \textbf{``Entropy-based''} baseline which selects the next best view based on entropy~\cite{sock2017multi, doumanoglou2016recovering} is presented to be able to compare to a recent evaluation. For this baseline, the pipeline is identical to the proposed system except the action is generated based on view entropy, which is pre-computed for every view. Segmentation masks inferred from the pose estimator are used for the view entropy calculation. Sample trajectories of each baseline are visualized in supplementary.

\noindent\textbf{Evaluation metrics:} Pose accuracy for all objects are evaluated with $e_{ADD}$ scores. For symmetric objects such as coffee cup, the hypothesis prediction corresponding to the axis of symmetry are ignored when $e_{ADD}$ was calculated. Following the standard practice~\cite{hinterstoisser2012model}, objects with $e_{ADD}$ error less than 10\% of the object diameter is considered to be a correct hypothesis. All pose errors are measured in mm. The detection rate is defined as the ratio of number of objects with correctly estimated pose to the total number of objects in the bin in each scenario. Distance traveled is the total distance the camera moved where it is measured along the surface of the hemisphere as it is unrealistic for the sensor to move through the objects.

\begin{figure}[t]
\centering
\includegraphics[width=\linewidth]{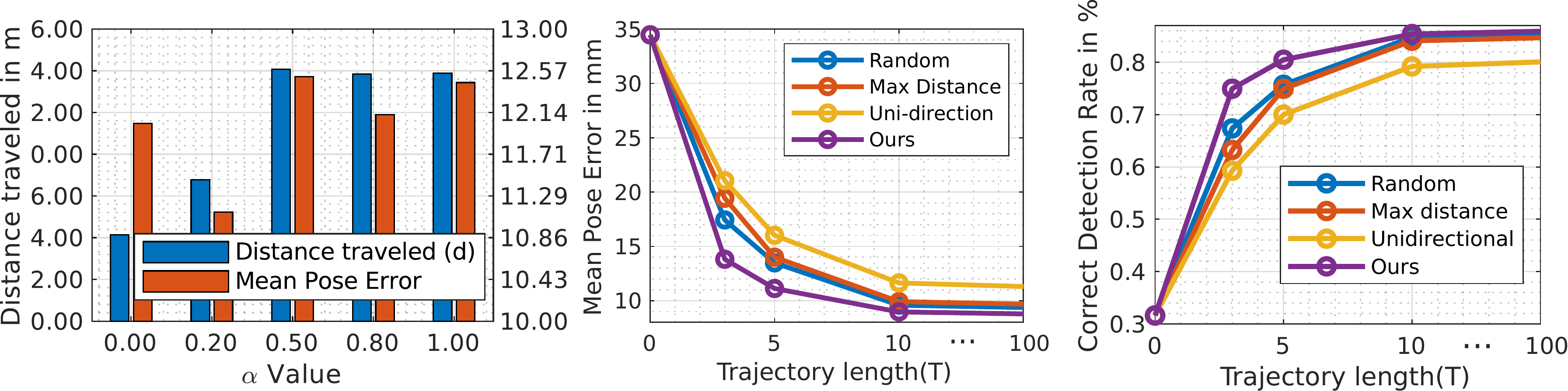}

\caption{Performances with different hyperparameters. (Left) Distance travelled and mean pose error with respect to $\alpha$ values. Change in mean pose error (Middle) and detection rate (Right) with different number of maximum episode length ($T$).}
\label{fig:Ablation_figures}
\end{figure}

\begin{figure}[t]
\centering
\includegraphics[width=\linewidth]{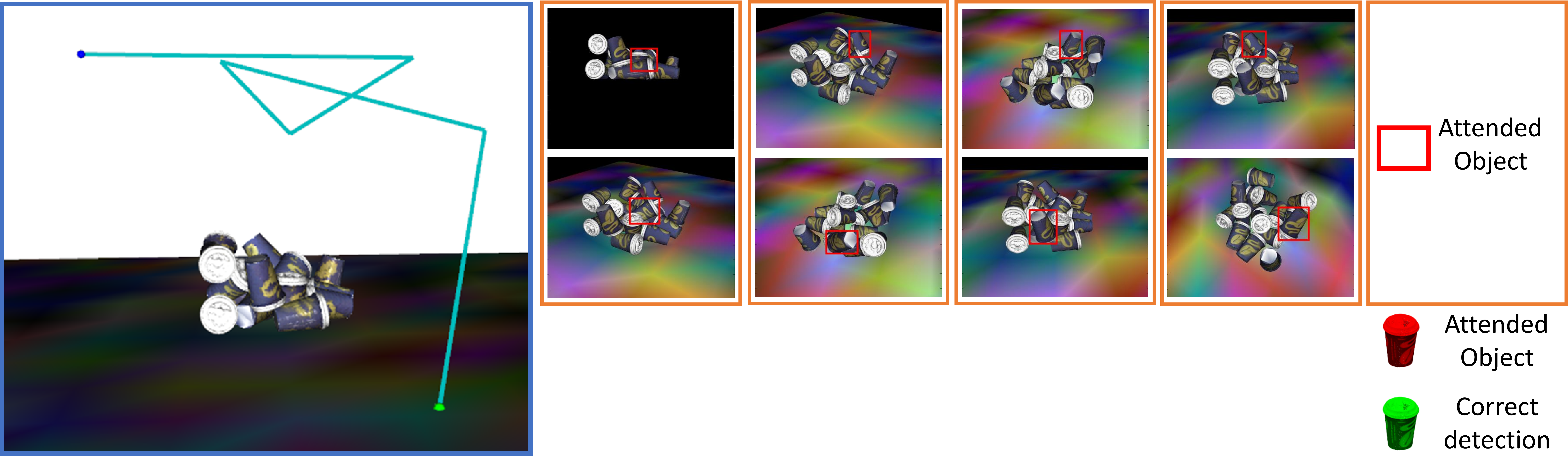}

\caption{(Left) An example of the learned trajectory for the coffee cup dataset. (Right) Every column represents one time step (Top and middle row) Image from the viewpoint before and after taking the action. (Bottom row) 3D visualization of the action for each time step.}

\label{fig:attention_visualized}
\end{figure}

\begin{figure}[t]
\centering
\includegraphics[width=0.9\textwidth]{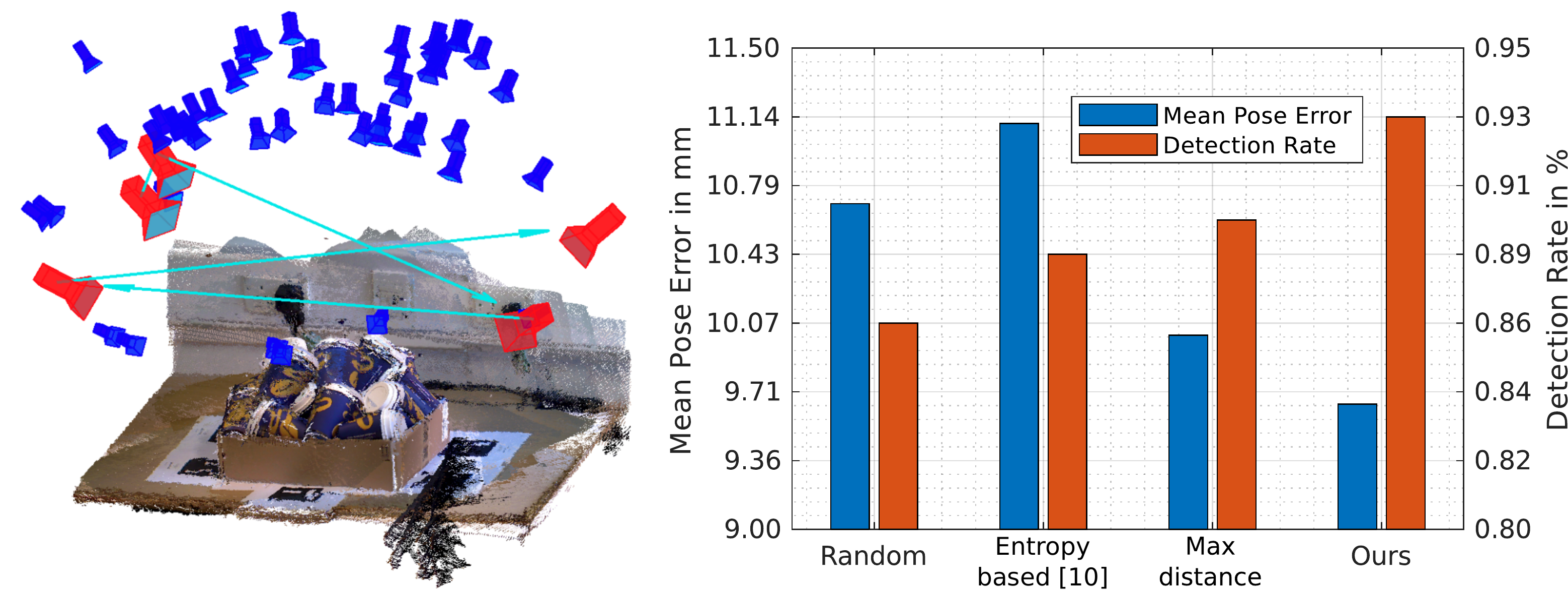}

\caption{Evaluation on real dataset. (Left) Visualization of a sample trajectory. Blue camera represents a set of allowable viewpoints and red cameras show the viewpoints selected by the proposed system. The arrow shows the order of view selection. (Right) Performance of different baselines on the real dataset.}
\label{fig:real_experiment_figures}
\end{figure}

\noindent\textbf{Synthetic dataset.} Table. ~\ref{tab:main_results} shows the mean pose error and the correct detection rate on the synthetic dataset. To obtain the results, episode lengths $T$ are held constant at 5 steps. For Coffee Cup object, our approach consistently outperforms all baselines in both $e_{ADD}$ and detection rate metric by a significant margin. In most cases, \textit{Unidirectional} policy results in low distance $d$ as the agent only moves horizontally. For Bunny object, the proposed method outperforms all baselines in all metrics with a narrower margin due to the smaller number of instances in the scene. Fig.\ref{fig:attention_visualized} shows how the agent behaves at each time step of the trajectory. The agent selects the object with the lowest confidence, which is expressed as the verification score, but also not too far away to reduce the distance traveled. The attention shifts to different objects in different time step and it can be noted the number of the correctly detected object (highlighted in green) increases every step.

\noindent\textbf{Real world dataset.} Fig.~\ref{fig:real_experiment_figures} shows a sample trajectory generated by the learned policy network and quantitative comparison with other baselines. Unidirectional baseline is not included since the distribution of viewpoints in the real dataset is not in a grid form or uniform. Since most of the viewpoints are densely populated on the top part of the scene, the learned policy tends to select views with lower elevation resulting in a behavior similar to maximum distance baselines. The proposed method outperforms the other baselines including the entropy-based method~\cite{sock2017multi}, showing it is important to choose views conditioned on the inference state. 

\subsection{Ablation experiments}
\noindent\textbf{$\alpha$ and $T$ parameters.} Experiment with different trajectory is presented to show how the performance changes within the spectrum between single-shot pose estimation ($T=0$) and all-view pose estimation ($T=99$ in our case). Fig.~\ref{fig:Ablation_figures} shows the performance indeed increases with more views but the benefit saturates after $T=10$. For $\alpha$, excessively emphasizing the $r^{dist}$ term results in the agent to give attention to the furthest object and move the camera towards the object to maximize the reward, increasing the travel distance. Whereas it is difficult for the agent to learn a policy without $r^{dist}$ term since during training often there are multiple views around the object of interest which are equally favorable.

\noindent\textbf{Attention module.} To verify if the attention module is properly learning to automatically select an object, a baseline which assigns attention to the object with the lowest verification score is tested. Compared to the verification score-based selection, our baseline improved the detection rate by 3\% and mean pose error by 9\%.\\


\section{Conclusion}
\label{sec:conclusion}
We presented a framework to deal with the active 6D object pose estimation problem in cluttered scenarios. We formulated our framework within a reinforcement learning and carefully designed all the different components achieving superior performance compared to different baselines. However, we believe there is a margin of improvement in terms of performance and real-world framework evaluation given the lack of proper datasets. As future work, we would like to explore the use of the proposed framework on a real robot, its impact on the graspability of objects and the use hand-object pose priors to accelerate the policy training~\cite{garcia2018first, antotsiou2018task}. Also, using a real robot could ease the real world dataset generation. 

\acknowledgments{ This work is part of Imperial College London-Samsung Research project, supported by Samsung Electronics.}


\bibliography{biblio}  
\clearpage
\input{supplementary.tex}
\end{document}

%% file: supplementary.tex
\appendix
\section*{Supplementary material}
\renewcommand{\thesubsection}{\Alph{subsection}}

\subsection{Verification score}
\begin{figure}[h!]
\centering
\includegraphics[width=0.9\linewidth]{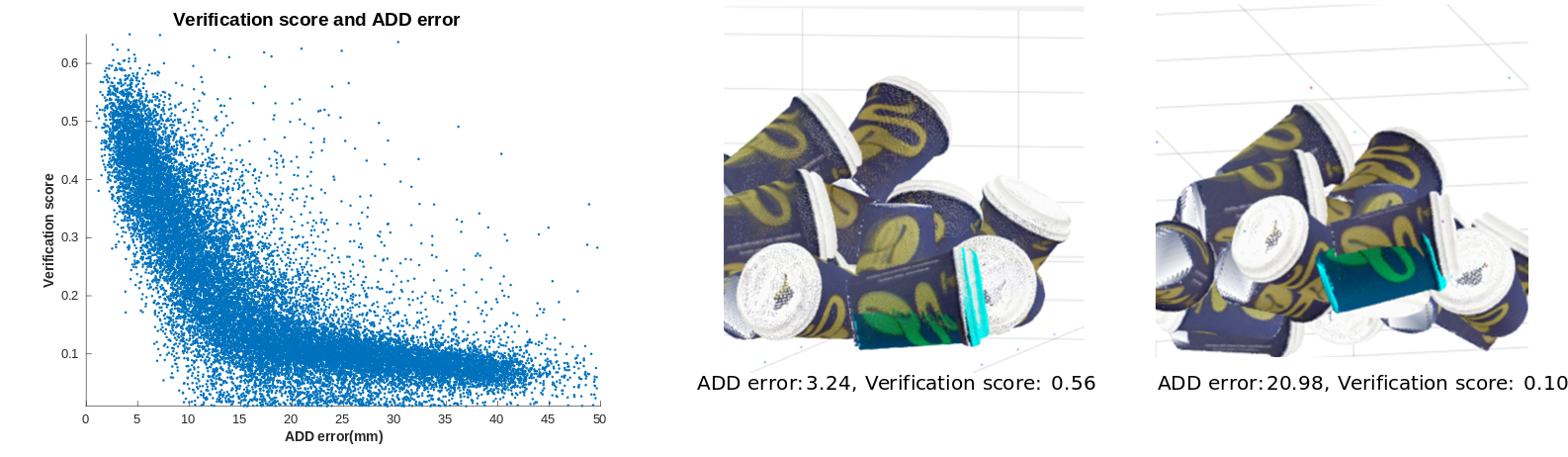}
\vspace{0.2em}
\caption{(Left) A graph showing a relation between verification score and $e_{ADD}$. (Middle and right) The figures show the scene and hypotheses rendered in blue.}
\label{fig:verification_score}
\vspace{-2.0em}
\end{figure}

In Fig.\ref{fig:verification_score}, a graph of verification score against the $e_{ADD}$ for coffee cup is presented showing how the proposed verification score correlates with $e_{ADD}$, making it suitable to be used in the absence of ground-truth annotations. For coffee cup object, 10\% of the diameter of the object is 14mm which is used to decide whether the object pose hypothesis is of an acceptable quality.

\subsection{Baselines}
\begin{figure}[t!]
\centering
\includegraphics[width=0.9\linewidth]{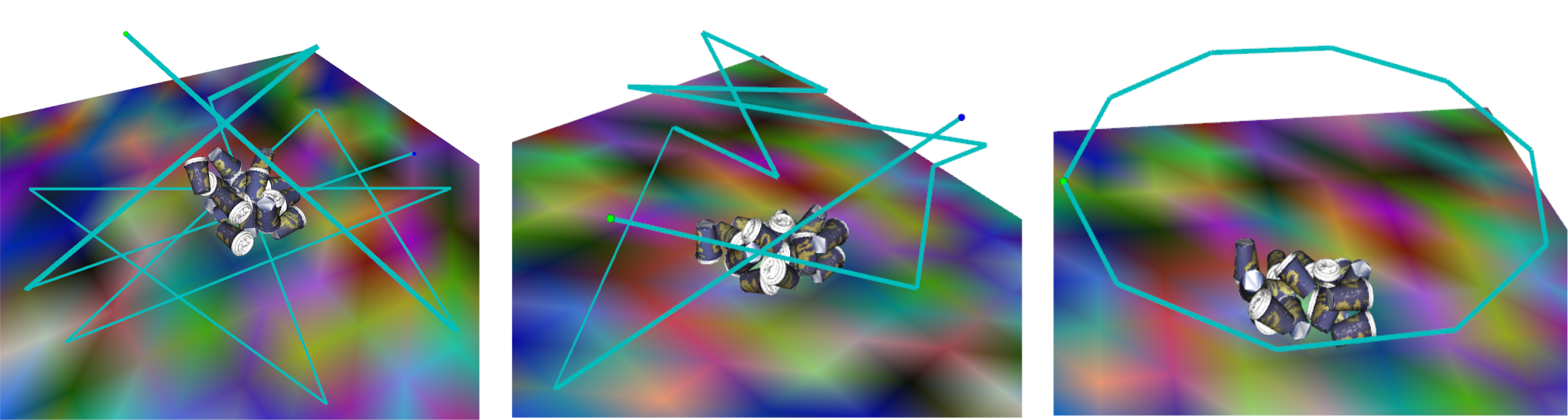}
\caption{(Left) Sample image of Maximum Distance policy baseline (Middle) Sample image of Random policy baseline (Right) Sample image of Unidirectional policy baseline }
\label{fig:baseline_trajectories}
\end{figure}

Sample trajectories of baselines are visualized in Fig.~\ref{fig:baseline_trajectories}. The first baseline \textbf{``Random''} samples random actions from a fixed Gaussian distribution. The second baseline \textbf{``Unidirectional'' } moves the camera in one horizontal direction around the scene. The trajectory is generated such that it completes one full revolution around the scene. Elevation of the viewpoint is fixed to be 45 degrees which provide a balanced view between occlusion and variability in each view. The third baseline, \textbf{``Maximum distance''}, selects viewpoints with the longest distance from previous viewpoints. This baseline test whether the most informative views are simply far-apart views.

\begin{figure}[t!]
\centering
\includegraphics[width=0.9\linewidth]{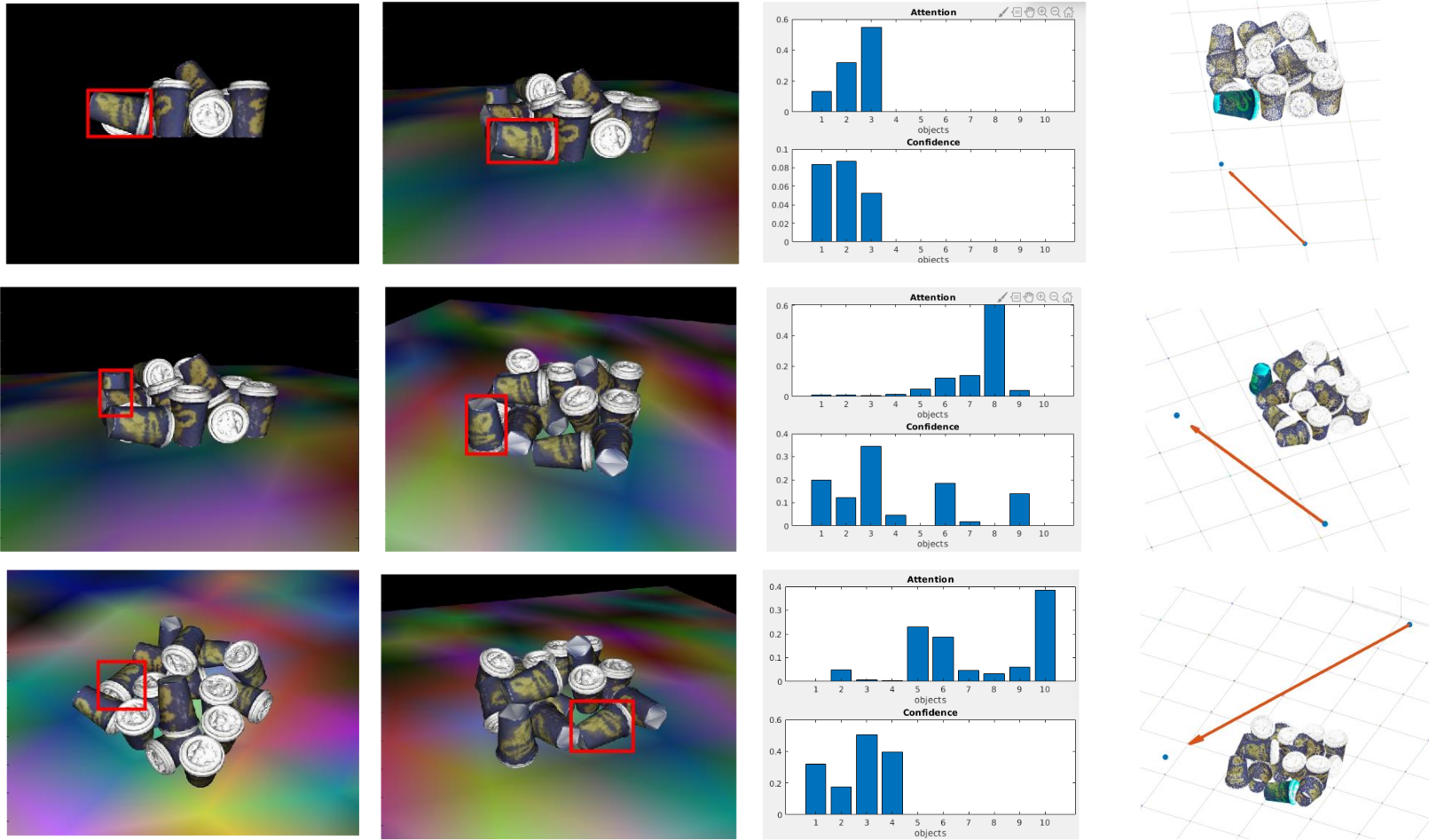}
\caption{Relationship between attention, verification score and action. (First column) Images of the scene before taking action. Attented object is highlighted in red box. (Second column) Images of the scene after taking action. (Third column) Bar graph showing the object score($\bar{w}^i$ in the manuscript) and verification score(shown as confidence). (Fourth column) 3D visualization of camera trajectory.}
\label{fig:qualitative_trajectories}
\end{figure}

\subsection{Qualitative evaluations}
More qualitative evaluation results including the attention histogram for detected object for better insights are shown in Fig.~\ref{fig:qualitative_trajectories}. It can be noted the selector network tends to choose the object with the most uncertainty.

\subsection{Dataset construction}
Each scene has 100 views on the upper hemisphere with radius of 80 cm. The view grid consists of 5 elevation and 20 azimuth levels and both RGB and depth images are rendered with OpenGL. The number of instances varies randomly between 15-20 for Coffee Cup and 7-12 for Bunny due to the size and shape difference. For $e_{ADD}$, all detectable objects are taken into account, and the undetected object are assigned the error of 50mm which is more than 3 times the acceptable threshold.

\subsection{Architecture Details}
$\it{Selector(\cdot)}$ is a linear mapping from object features to attention score which is optimized jointly with the policy. The module maps 12 dimensional input to a single scalar value for each object. The policy network has two fully connected layers where the first fully connected layer maps the features to 128 dimensional vector followed by ReLU activation function. The second fully connected layer maps to the action output. The list of hyper-parameters are shown in Table~\ref{tab:hyperparam}.

\begin{table}[]
\centering
\caption{List of the hyper-parameters}
\label{tab:hyperparam}
\begin{tabular}{|c|c|c|}
\hline
Hyper-parameter   & Value & Description     \\ \hline
Gamma             & 0.995 & Discount factor \\ \hline
Learning rate     & 1e-4  & Adam optimizer  \\ \hline
Number of batches & 128   &                 \\ \hline
Iteration size    & 10e6  &                 \\ \hline
\end{tabular}
\end{table}